# Automated Tomato Maturity Estimation Using an Optimized Residual Model with Pruning and Quantization Techniques

Muhammad Waseem[1], Chung-Hsuan Huang[2], Muhammad Muzzammil Sajjad[3], Laraib Haider Naqvi[4], Yaqoob Majeed[5*], Tanzeel Ur Rehman[1], Tayyaba Nadeem[6]

[1]Department of Biosystems Engineering, Auburn University, Alabama, 36890, USA
[2]Department of Electronics and Electrical Engineering, National Yang-Ming Chiao Tung University, No. 1001, Daxue Rd, East District, Hsinchu City, Taiwan, Republic of China.
[3]Department of Food Engineering, University of Agriculture, Faisalabad, Pakistan
[4]Department of Mechanical Engineering, University of Georgia, Athens USA.
[5]Department of Electrical Engineering and Computer Science, University of Wyoming, USA.
[6]Department of Food Technology, National Institute of Food Science and Technology, University of Agriculture, Faisalabad, Pakistan.
* Corresponding Author: ymajeed@uwyo.edu (Yaqoob Majeed)

**Abstract**

Tomato maturity plays a pivotal role in optimizing harvest timing and ensuring product quality, but current methods struggle to achieve high accuracy along computational efficiency simultaneously. Existing deep learning approaches, while accurate, are often too computationally demanding for practical use in resource-constrained agricultural settings. In contrast, simpler techniques fail to capture the nuanced features needed for precise classification. This study aims to develop a computationally efficient tomato classification model using the ResNet-18 architecture optimized through transfer learning, pruning, and quantization techniques. Our objective is to address the dual challenge of maintaining high accuracy while enabling real-time performance on low-power edge devices. Then, these models were deployed on an edge device to investigate their performance for tomato maturity classification. The quantized model achieved an accuracy of 97.81%, with an average classification time of 0.000975 seconds per image. The pruned and auto-tuned model also demonstrated significant improvements in deployment metrics, further highlighting the benefits of optimization techniques. These results underscore the potential for a balanced solution that meets the accuracy and efficiency demands of modern agricultural production, paving the way for practical, real-world deployment in resource-limited environments.

## 1. Introduction

Tomatoes (*Solanum lycopersicum*), a linchpin in global agricultural production, necessitate a comprehensive understanding and application of maturity indices. These indices-ranging from Green, Breaker, Pink, Light Red, Red, and Over mature-are essential for optimal cultivation and distribution (Dinu et al., 2023; Khatun et al., 2023). These indices, pivotal in determining harvest timing, significantly impact the produce's quality, nutritional value, and economic returns (Shinoda et al., 2023). The stage of maturity influences not only the organoleptic properties of tomatoes such as flavor and texture, but also their processing suitability and shelf life (Choi & Park, 2023; Shezi et al., 2023). Accurate maturity assessment minimizes wastage, optimizes logistics, and ensures a consistent supply of high-quality tomatoes, a critical factor in addressing the nutritional needs of the world's growing population (Du et al., 2023; Rusu et al., 2023).

The integration of automated classification systems such as AI-driven maturity indices based on image processing and machine learning-based quality assessments in agriculture signifies a revolutionary shift in crop management and quality control strategies (Wakchaure et al., 2023). These AI-driven approaches provide scalable and objective alternatives to manual assessments, enhancing consistency and reducing human error. Moreover, while deep learning techniques have improved classification accuracy, these models suffer from high computational demands that limit their scalability in resource-constrained environments (Filip et al., 2020; Jha et al., 2019). These technologies, provide the potential to make significant advancements in the postharvest handling of fruits and vegetables (Majeed & Waseem, 2022). Incorporating such systems empowers farmers and distributors with data-driven insights, fostering increased productivity and sustainable agricultural practices.

Mputu et al. (2024) introduced ML algorithms for tomato quality sorting, achieving a classification accuracy of 89% while it does not fully address computational efficiency issues for real-time applications. Qasrawi et al. (2021) achieved an accuracy of 70.3% with neural networks and 68.9% with logistic regression models for tomato classification, particularly mature-stage tomatoes with diseases. Vo et al. (2024) applied YOLOv9 for automating tomato ripeness classification (Unripe, Semi Ripe, Fully Ripe), achieving a precision of 0.856, recall of 0.832, and mAP50 of 0.882. Nath and Roy (2023) reached 80% accuracy using CNN for sentiment analysis to classify text into positive, negative, or neutral sentiments of rotten tomatoes. Zhang et al. (2024) developed a YOLOv5-based algorithm for tomato detection and pose estimation achieving 82.4% accuracy on un-occluded tomatoes for harvesting. Quach et al. (2024) used MobileNet models for Unripe, ripe, Old, and damaged tomato classification and visual explanations. Chen et al. (2021) worked on the tomatoes classification system for green, breaker, turning, pink, light red, and red stages achieving an accuracy of 81.3% from Denesenet, 83.5% from ResNet, 69.1% from MobileNet, 67.4% from AlexNet, 58.3% from Shufflenet, 85.6% from SqueezeNet based on open source dataset. Pacheco and López (2019) worked on tomato classification for green, breaker, turning, pink, light red, and red stages using K-NN and MLP neural networks based on color, achieving an accuracy of 90%. The heavy reliance on color-based features restrict the method's robustness under different environmental conditions. de Oliveira et al. (2023) achieved 92% accuracy in classifying organic versus non-organic tomatoes using Decision Trees on mass spectrometry data. Despite the high accuracy, the approach is less practical for large-scale or real-time applications due to its dependence on expensive and time-consuming mass spectrometry. Garcia et al. (2019) automated ripeness classification of tomatoes, the study's reliance on SVM yielded 83.39% accuracy on a limited dataset of 900 images, restricting its effectiveness and scalability in real-world applications. The above-mentioned studies faced limitations related to scalability, accuracy, and practicality. Few of the studies relied on limited open-source datasets, making them less effective in real-world applications. Machine learning models such as SVM and logistic regression, which use hand-crafted features, further restricted the generalizability of these approaches. Additionally, while deep learning-based methods offer greater accuracy, they are computationally intensive, demanding substantial processing power, which limits their practical adoption in resource-constrained environments. Therefore, it is essential to develop models that balance classification accuracy with computational and energy efficiency for real-world deployment. In recent years, techniques such as pruning and quantization have been introduced to enhance the computational efficiency of deep learning models. Pruning helps to reduce the size of a neural network by removing less significant weights, allowing the model to maintain accuracy while using fewer resources. On the other hand, quantization reduces the precision of the model's parameters, such as converting 32-bit floating-point numbers to 8-bit integers, which lowers the computational and memory requirements without significantly sacrificing performance. These approaches are especially beneficial for deploying models in resource-constrained environments.

Therefore, the primary goal of this research is to develop a computationally efficient tomato classification model using the ResNet-18 architecture. This study seeks to overcome the challenges posed by computationally intensive deep learning networks using pruning and quantization approaches, enabling their deployment in resource-constrained agricultural environments. The first objective of this research is to optimize ResNet-based model to accurately classify tomatoes based on maturity stages. The second objective is to make the ResNet model lighter and easier to deploy on edge devices by leveraging pruning and quantization approaches. The final objective is to deploy the optimized model on the edge device to investigate its performance capabilities for real-time classification of tomato maturity. The results from this study would demonstrate the deep learning model's real-time utility, showcasing its ability to operate efficiently on a low-power edge device.

## 2. Material & Methodology

The methodology employed in this work, as illustrated in Figure 1, was comprised of data collection using Raspberry color camera V2.1, deep learning model training along with optimization and quantization, and model deployment and testing on the Jetson platform.

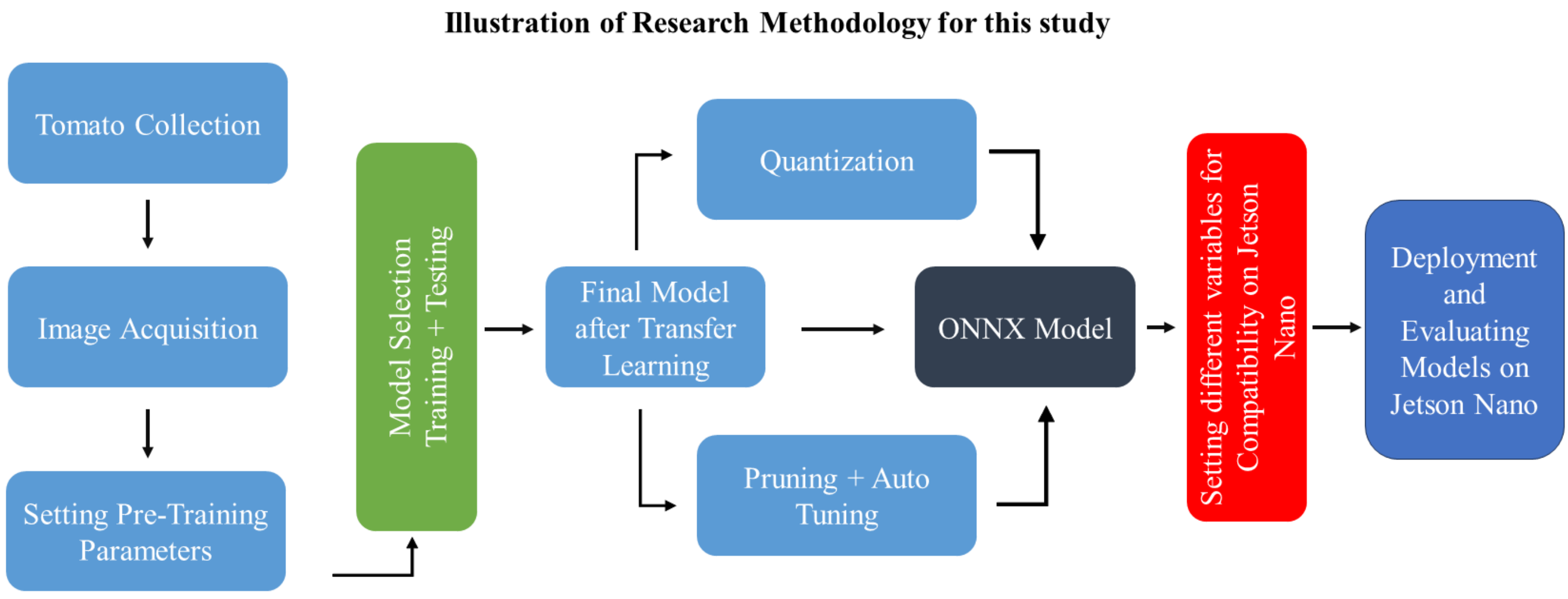

Figure 1: Illustration of research methodology for this study.

### 2.1 Image Acquisition:

To collect the image dataset for this study, *Solanum lycopersicum* tomato (Rio Grande) variety was procured from the local market. A total of five vendors were randomly selected to collect the tomato samples to ensure the variability of the dataset. A total of 25 tomato samples were chosen for each maturity stage (Green, Breaker, Pink, Light Red, Red, and Over mature) as per USDA standards (USDA, 1991) with consideration provided by Serge Ndayitabi (2020) and (Garcia et al., 2019) making up to a total of 150 tomato samples, as given in Table 1. The selection criteria for the tomatoes were based on USDA standards: Green (completely green), Breaker (10% or less color break), Pink (30%-60% pink/red), Light Red (60%-90% red), Red (over 90% red), and Over Mature (signs of over-ripeness). To ensure accurate classification, each sample was verified by a researcher assessed in a controlled environment with consistent lighting, documented with high-resolution images, and classified by trained personnel from Horticulture department of University of Agriculture, Faisalabad. While direct color evaluation based on sensory panel assessments is a well-established method for determining tomato maturity, our approach leverages a deep learning classification network to capture a broader range of features. In addition to color, the network learns texture, shape, and subtle morphological characteristics that are especially important when color differences are marginal. Furthermore, by training on expert sensory evaluation data, the network internalizes the intricate distinctions that experts use to classify the tomatoes. This results in a model that is robust to variations in illumination and imaging conditions, thereby offering a more comprehensive and reliable solution compared to isolated color analysis. Figure 2 shows the example tomato image from each class. This classification scheme was designed to capture the nuanced changes in color and ripeness as the tomatoes transitioned through different stages, facilitating the development and validation of a robust maturity indices system, as proposed for this study. To ensure the reliability of the dataset and to avoid data leakage, the splitting of data into training and test sets was performed at the tomato-sample level rather than at the image level. That is, all 12 images from a single tomato (captured from six orientations under two lighting conditions) were assigned exclusively to either the training set or the test set. This strategy prevents the overlap of highly similar images between sets and ensures that the model's generalization is tested on truly independent samples.

Table 1: Overview of the complete dataset of tomatoes captured under different conditions from different angles.

| Tomato Class | Total Tomatoes | Image captured for each tomato | Condition (On/Off) | Total Images | Overall dataset |
|---|---|---|---|---|---|
| Green | 25 | 6 | Light | 150 +150 | 1800 |
| | | | Without Light | | |
| Breaker | 25 | 6 | Light | 150 +150 | |
| | | | Without Light | | |
| Pink | 25 | 6 | Light | 150 +150 | |
| | | | Without Light | | |
| Light Red | 25 | 6 | Light | 150 +150 | |
| | | | Without Light | | |
| Red | 25 | 6 | Light | 150 +150 | |
| | | | Without Light | | |
| Over Mature | 25 | 6 | Light | 150 +150 | |
| | | | Without Light | | |

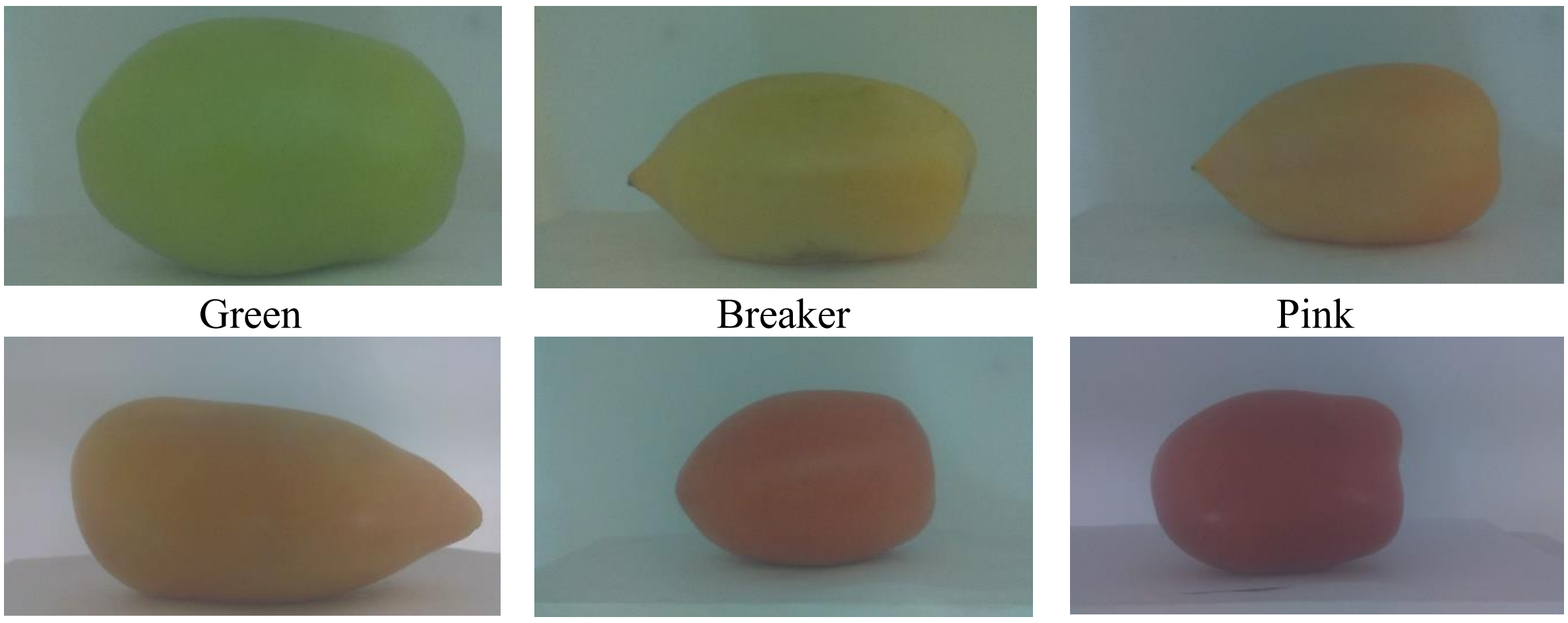

Green Breaker Pink

Light Red Red Over Mature

Figure 2: Different stages of the tomato's dataset

To acquire the images of tomatoes, an image acquisition system was developed, as shown in Figure 3 (a). This image acquisition system mainly consisted of an imaging chamber, Raspberry Pi Camera v2.1, and NVIDIA Jetson board. The imaging chamber was equipped with a light source to control the illumination. Placing the tomato within this chamber allows for the simulation of different lighting conditions, capturing the tomato's images under both ambient and controlled illumination. This enhances the robustness of the dataset, making it versatile and adaptable to diverse environments. Raspberry Pi Camera v2.1 was integrated with NVIDIA Jetson board through the Camera Serial Interface (CSI) to collect the images. Raspberry Pi Camera v2.1 has an 8MP resolution of 3280 x 2464 pixels. The image sensor used in this camera was Sony IMX219 which can capture images at 4k resolution. The video capacity of this camera is 1080p, 30fps resolution.  While NVIDIA Jetson nano board was used as a central processing unit. The Jetson Nano features a quad-core ARM Cortex-A57 processor, an onboard 128-core Maxwell GPU with 2GB memory. An overview of the image acquisition system used in our research is illustrated in Figure 3 (b). Each tomato was captured from six distinct orientations: frontal, dorsal (back), up, down, left lateral, and right lateral (Figure 4), encompassing potential variances in skin texture, color, and other morphological features that are critical for accurate classification. The images were captured under two distinct lighting conditions (with and without artificial light). The first set of images was acquired in the absence of any external light source, simulating conditions where natural light prevails. The second set involved capturing images with a designated light source, replicating scenarios where additional illumination might influence the visual characteristics of the tomatoes. In total, each tomato underwent the capture of 12 (6 sides, each side in two illumination conditions i.e., Light source turned ON and light source turned OFF) images, resulting in a dataset comprising a total of 1800 images.

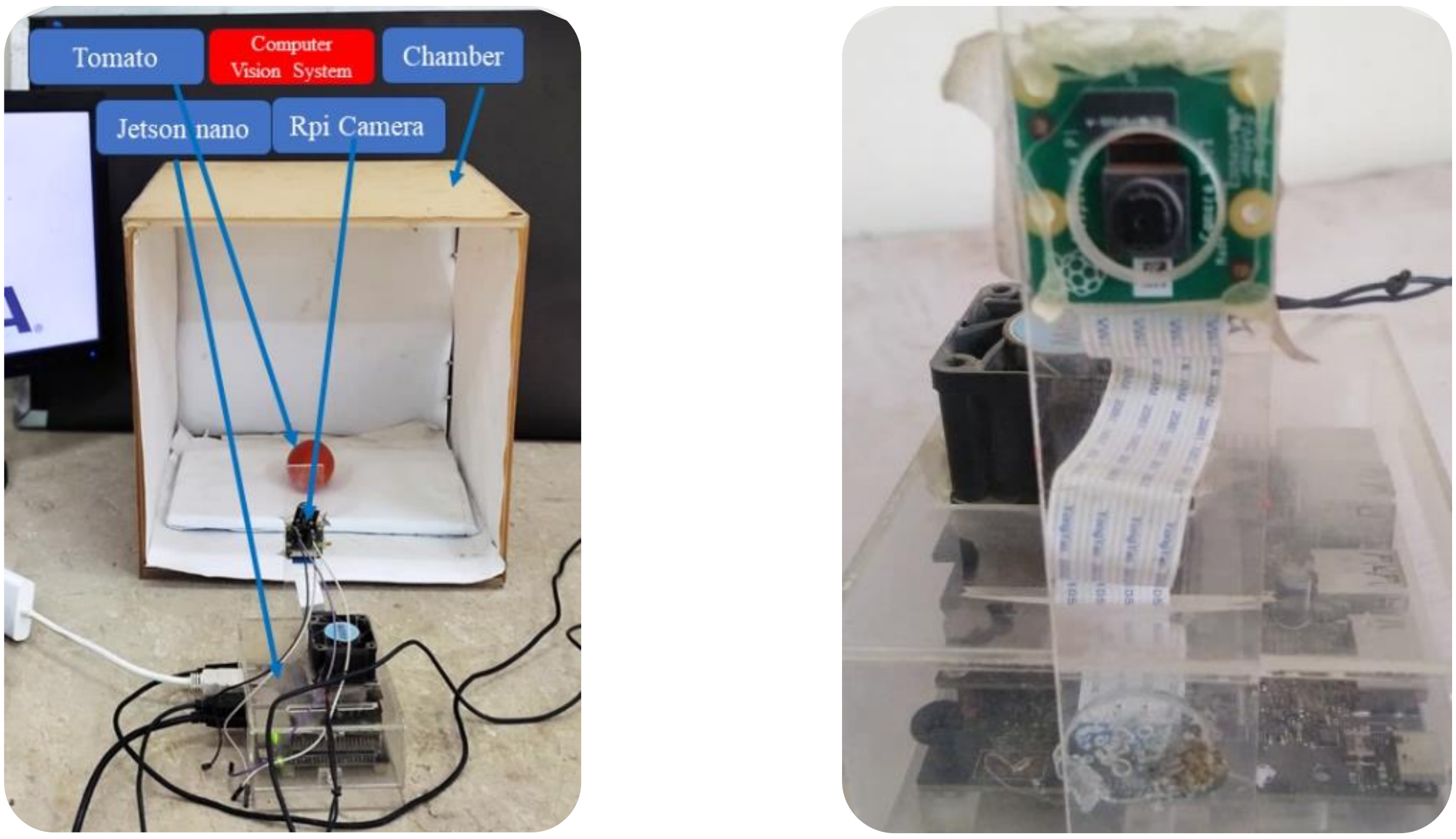

(a) (b)

Figure 3: Computer vision system for this study (a), Front view of integrated RPi camera via CSI with Jetson Nano (b).

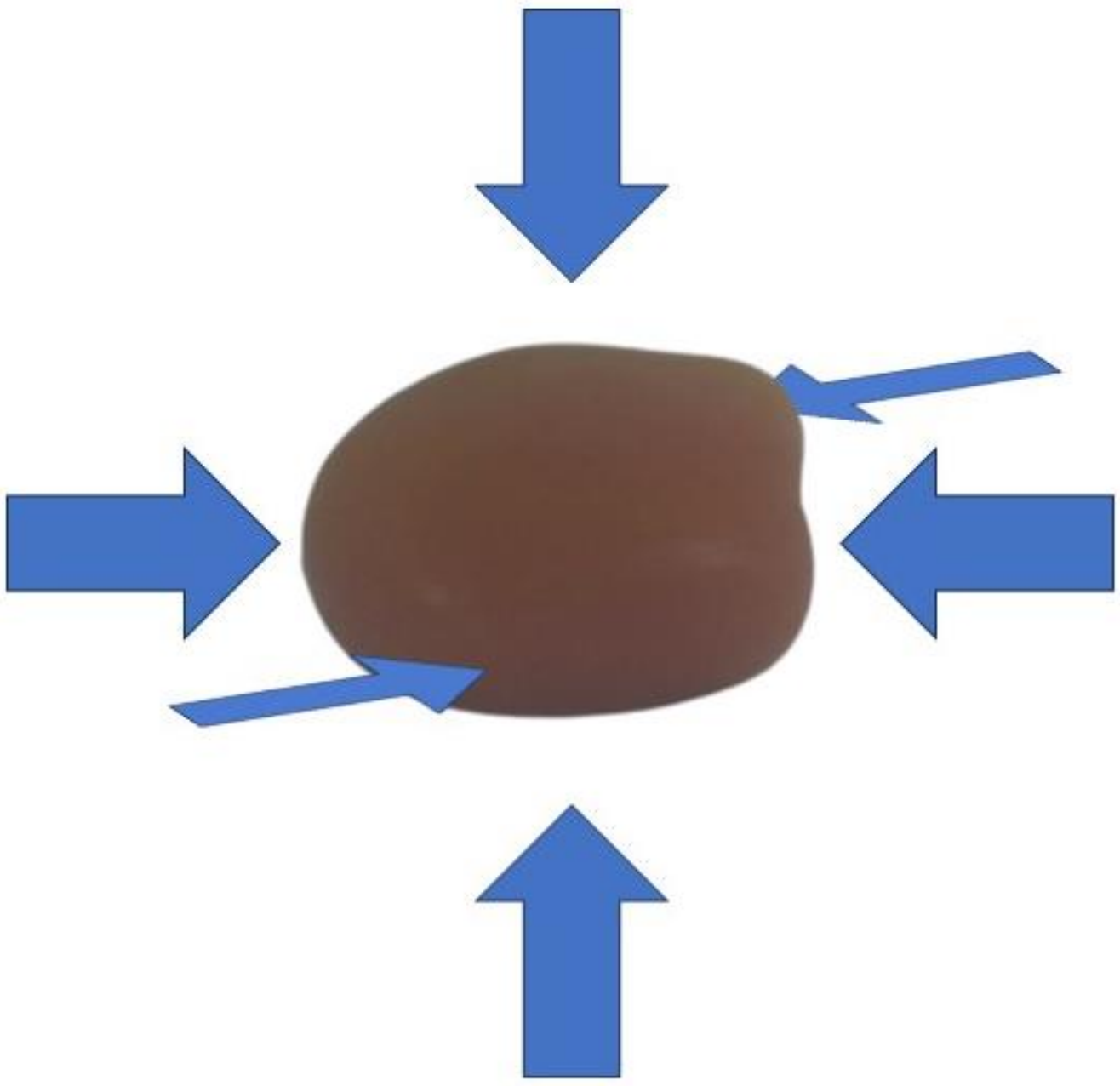

Figure 4: Illustration of orientation used to capture tomato images

### 2.2 Model Architecture

Residual networks, particularly ResNet18 (He et al., 2016), was chosen for this study due to its favorable balance between accuracy and computational efficiency. Its architecture, characterized by the use of skip connections, enables effective training by alleviating the vanishing gradient problem, thus ensuring robust feature extraction even in deeper networks. Moreover, the relatively lightweight nature of ResNet-18 makes it well-suited for deployment on resource-constrained edge devices an important consideration for real-time agricultural applications. ResNet18 consists of eighteen layers that utilize these skip connections shown in Figure 5. The architecture and building blocks of ResNet18, as illustrated in Table 2, form the foundation of this study due to their robustness and high performance.

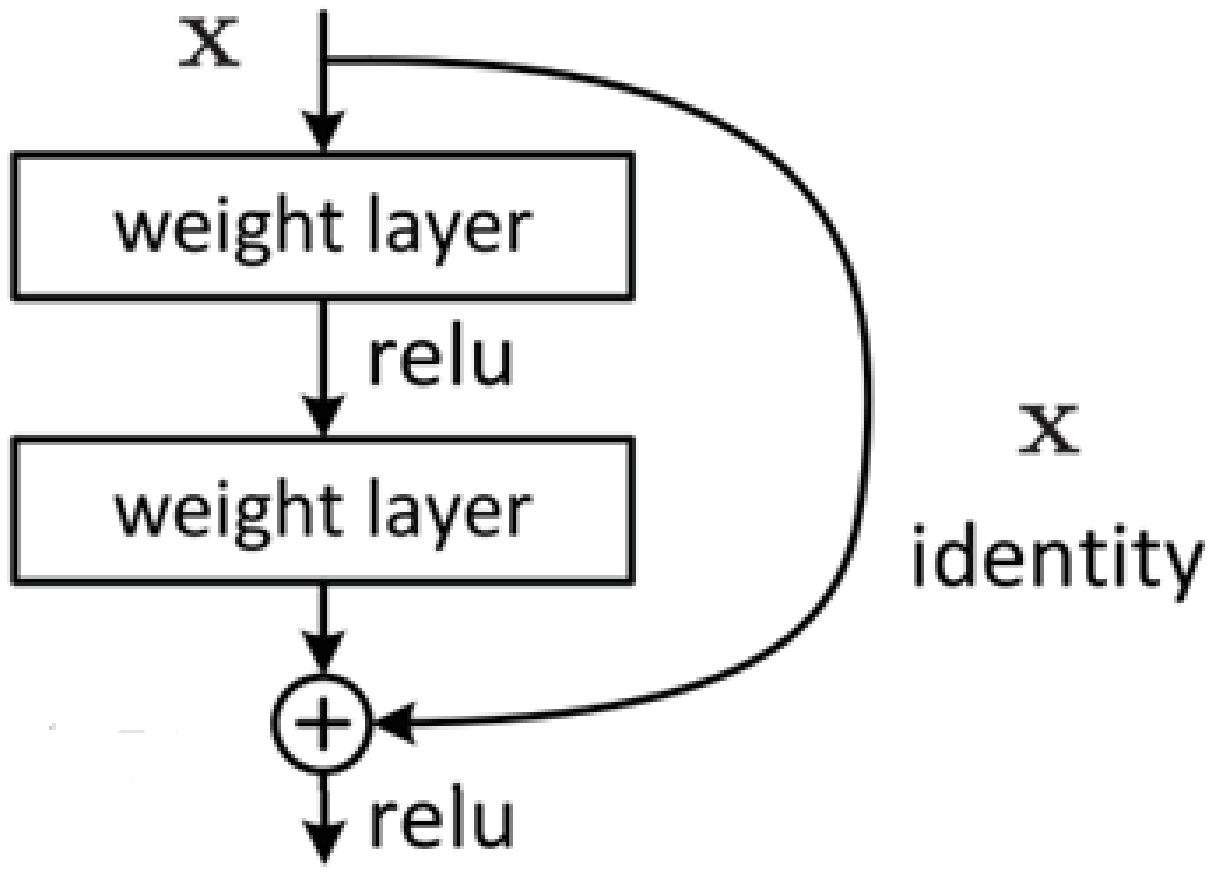

Figure 5: Illustration of skip connection in a single building block of ResNet model

Table 2: Architecture and building blocks of base ResNet-18 used in this study

| Layer Name | conv1 | conv2 x | | conv3 x | conv4 x | conv5 x | | FLOPs |
|---|---|---|---|---|---|---|---|---|
| Output Size | 112 x112 | 56 x 56 | | 28 x 28 | 14 x 14 | 7 x7 | 1 x1 | |
| 18-Layer | 7 x 7, 64, stride 2 | 3 x 3 max pool, stride 2 | 3 x 3, 64<br>3 x 3, 64 | 3 x 3, 128<br>3 x 3, 128<br>x 2 | 3 x 3, 256<br>3 x 3, 256<br>x 2 | 3 x 3, 512<br>3 x 3, 512<br>x 2 | Average pool, softmax, fc | $1.8 \times 10^9$ |

ResNet18 consists of 18 layers, including a 7x7 kernel as the first layer. It comprises four layers of identical ConvNets. Each layer has 2 residual blocks. Each block consists of two weight layers, with a skip connection linked to the output of the second weight layer via a ReLU. ResNet18 has an input size of (224,224,3). The fully Connected (FC) layer is the ultimate output (Limonova et al., 2021). The base ResNet architecture is shown in Figure 6, by modifying the final fully connected layer to output six classes, we tailored the network to capture the unique features of each tomato class. This customization enabled the model to achieve high accuracy across all six categories, demonstrating its effectiveness in precise and comprehensive tomato classification. We retained the original 7×7 convolutional layer with stride 2 and padding 3 for the input size (224, 224, 3). The convolutional and residual blocks were pruned and quantized to reduce parameters, and we replaced the original fully connected layer (1,000 classes) with a new one for six classes.

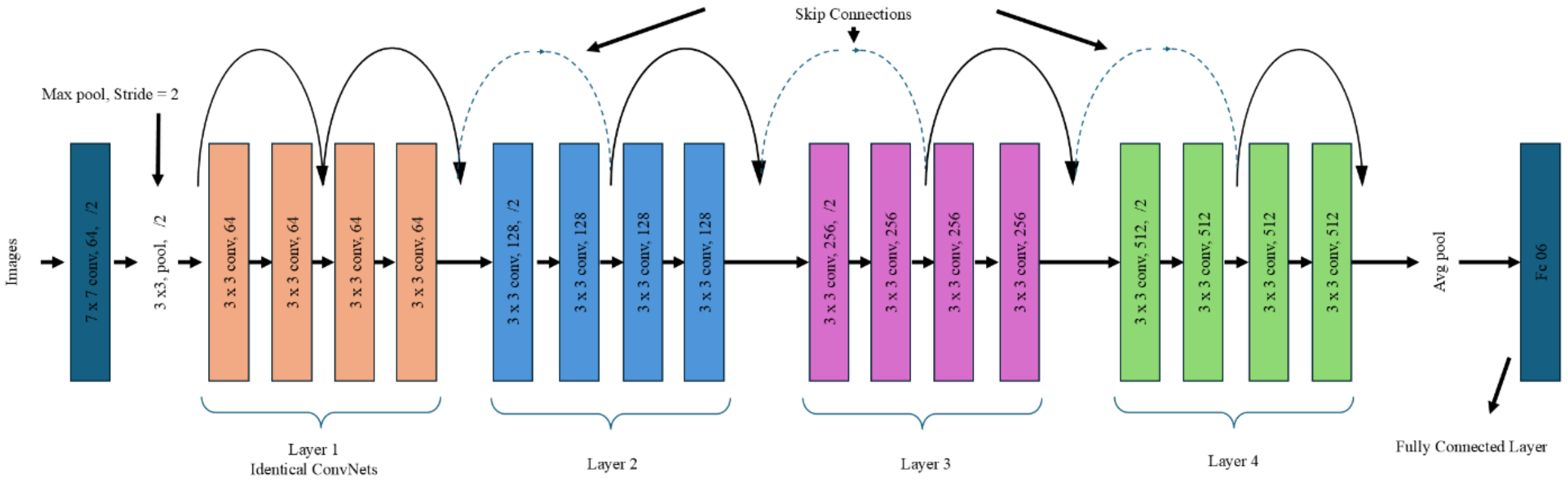

Figure 6: Detailed overview of the residual network with 18 layers used in this study, highlighting convolutional layers, and building blocks that facilitate deep learning by enabling information

### 2.3 Pruning and Auto-tuning

Transfer learning with deep networks like ResNet-18 achieve state-of-the-art accuracy, its heavy weight architecture with a large number of parameters and layers requires substantial computational resources during inference. This leads to increased time, power, and memory consumption, making the inference phase expensive and less suitable for deployment on resource-constrained devices (Molchanov et al., 2016). Pruning could help in reducing network complexity by eliminating redundant parameters and connections, thereby decreasing the model's size and computational requirements. Pruning a CNN model involves decreasing the model's parameters, although this might result in a decline in the model's performance (Poyatos et al., 2023).

Following the application of transfer learning using ResNet-18, modified to accommodate the six target tomato classes by replacing the original fully connected layer. we implemented magnitude pruning to enhance model efficiency for real-world applications. Magnitude pruning is a model compression technique that eliminates less significant weights based on their magnitudes, thereby reducing computational and memory burdens while maintaining model performance. Specifically, we applied L1 unstructured pruning to the convolutional and linear layers of the network. For each weight $\theta_i$ in the set of network parameters are given in Equation (i).

$$\theta = \{\theta_i\}^{n}{}_{i=1} \tag{i}$$

we calculated its absolute value $|\theta_i|$ and pruned weights (Equation (ii)) that satisfied $|\theta_i|<\tau$ where $\tau$ is a threshold determined using quantiles to retain only weights above the 67th percentile:

$$\theta_i^{pruned} = \{\, 0\, if\ |\theta_i| < \tau, \theta_i\ if\ |\theta_i| \geq \tau. \} \tag{ii}$$

This process resulted in a significant parameter reduction of 66.92%, decreasing the total number of parameters from 11,179,590 to 3,697,762. Notably, the pruning preserved the structural integrity of ResNet-18's core components, including convolutional layers {Conv2d}, batch normalization layers {BatchNorm2d}, ReLU activations, and {BasicBlock} structures. This preservation ensured that the essential architecture remained intact, allowing the pruned model to maintain its effectiveness. The pruning effectively reduced parameters that contributed minimally to the output, satisfying the condition $L(\theta^{\text{pruned}}) \approx L(\theta)$ for the loss function $L$. The overall methodology is illustrated in Figure 7. After pruning, additional fine-tuning is necessary because pruning alters the network's architecture by removing weights or neurons, which can disrupt the learned representations and lead to a decrease in model performance. Fine-tuning allows the remaining weights to adjust and compensate for the pruned connections, helping the network to recover lost accuracy by re-optimizing the parameters within the new, sparser architecture. The process of retraining the CNN on a custom or targeted dataset after pruning is called auto-tuning or fine-tuning.

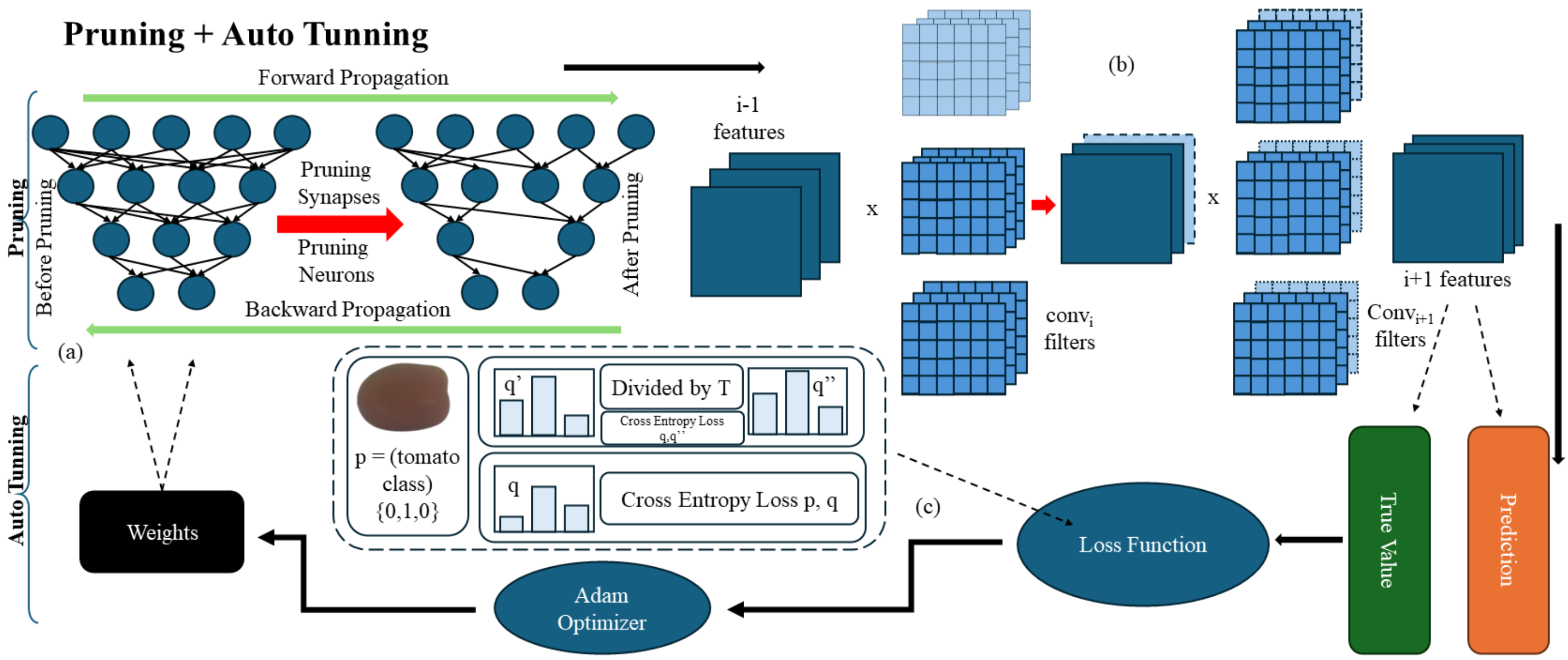

Figure 7: A deep insight into the neural network pruning and auto-tuning process. (a) Pruning while forward and backward propagation, (b) Filters before and after pruning for more efficient feature selection, (c) The cross-entropy loss function by the Adam optimizer to update weights.

The process of pruning is shown in the above horizontal row of figure 7, and the below row follows auto-tuning, and their interlink is illustrated. The whole process is divided further into three sections as (a), (b), and (c). The section (a) illustrates unstructured pruning which involves the removal of individual synapses, or connections, in a neural network without any specific pattern or structure. This was done based on the magnitude of the weights. The identification and elimination of connections with low-importance was made to reduce the overall size of the network. In this study, a simple feedforward network with multiple layers was used. Each connection between neurons had a corresponding weight. During unstructured pruning, we analyzed the weights of these connections, and connections with weights below a 66% threshold are pruned, effectively removed them. Section (b) illustrates the workflow of pruning L1-unstructured with the context of feature map removal. In the initialization phase of this study, a neural network model undergoes training using standard techniques without any pruning, featuring multiple layers with filters or neurons. Following this, the L1 of each parameter within the network is computed, representing the sum of the absolute values of its elements, where $i$ ranges from 1 to L. $W^{(i)}$ represents the connection weight matrix for the $i$-th layer shown in Equation (iii), with M$i$ as the number of input channels, N$i$ as the number of output channels, and K × K as the kernel dimensions.

$$\{W^{(i)} \in R^{(Mi\ x\ Ni\ x\ K\ x\ K)}, 1 \leq i \leq L\} \quad \text{(iii)}$$

Parameters possessing smaller L1-norm values, indicative of lesser importance to overall model performance, were identified and earmarked for potential pruning. Subsequently, parameters with L1 values falling below 66% threshold undergo pruning, entailing the removal of entire filters or channels in convolutional layers. This pruning process induces a structural change, necessitating adjustments in the filters of the subsequent convolutional layer to maintain network architecture and connectivity. The adjustment involved redistributing weights or reassigning connections. This was done by fine-tuning to further enhance or restore its performance. This iterative process of L1based model pruning ensures network size reduction while preserving or improving its functionality.

Section (c) illustrates the auto-tuning or fine-tuning of the pruned model, particularly focused on minimizing the loss and increasing accuracy. In this study, a ResNet-18 model previously pruned through magnitude pruning was loaded. The dataset is then split into training and validation sets, and the model is trained using the training set while monitoring its performance on the validation set. The Adam optimizer is employed with a learning rate of 0.001, and the CrossEntropyLoss is used as the loss function. The training is performed over several epochs, and for each epoch, both training and validation losses and accuracy were monitored. The fine-tuned model is subsequently saved. Fine-tuning allowed the pruned model to adapt to the specifics of the new dataset, potentially enhancing its performance compared to the original pruned model. In this study, the loss function is minimized as the input image is labeled as belonging to the Tomato maturity class, with the probability expressed as $\{0, 1, 0\}$. Following the inference of both the pruned and fine-tuned networks, the algorithms produce classification results q and q'. The image exhibits traits associated with close maturity class that are not distinctly reflected in q and q'. emerges, resulting in a classification result q'' that provides more intricate insights. By training the pruned network with fine-tuning, the final network enhances its accuracy based on the distinctive characteristics acquired during the pruning process.

### 2.4 Quantization

In contrast to pruning, where less significant weights are removed to reduce the number of parameters in a neural network, quantization focuses on reducing the precision of the network's weights and activations to lower bit widths also helping to decrease memory and computational requirements. In this study, during the initial configuration phase, the classification task is defined with 6 classes, with a specified directory for saving the quantized model. The pre-trained ResNet-18 model was loaded without its final fully connected layer, which was then adjusted to accommodate the new task's 6 classes. The model was loaded with weights from a previously trained model (transfer learning), set to evaluation mode, and path for test data and data transformations were defined. The test dataset was loaded and transformed using **torchvision's ImageFolder**, and a **DataLoader** were created for efficient batching. Quantization and dequantization stubs (**QuantStub** and **DeQuantStub**) were introduced to mark regions for quantization as illustrated in Figure 8. Quantization and dequantization are essential stages in converting a model's weights and activations from floating-point precision to lower bit precision, like int8, aiming to reduce memory storage demands and enhance computational efficiency during inference. In addition to reducing the memory and computational demands of the model, quantization yields improvements in classification accuracy. One reason for this improvement is that quantization introduces a form of regularization by limiting the precision of the model's weights and activations. This constraint prevents the model from overfitting to noise in the training data, thereby enhancing generalization performance. Moreover, the nature of our image dataset where images are captured under lighting and from multiple orientations ensures that the essential features are retained even when parameter precision is reduced. This enables the quantized model to effectively extract and utilize discriminative features, contributing to both its efficiency and accuracy. The QuantStub identified the input tensor region where quantization was applied, serving as the starting point for quantizing the input Equation iv, v vi respectively. X be the input tensor to the model, Q(X) be the quantized tensor, and D(Q(X)) be the dequantized tensor.

$$Q(X) = QuantStub(X) \quad \text{(iv)}$$

On the other hand, the DeQuantStub marked the output tensor region where dequantization occurred, signifying the end of the quantized region where the tensor is converted back to floating-point precision.

$$D(Q(X)) = DeQuantStub(Q(X)) \quad \text{(v)}$$

The quantization configuration is set using the **'fbgemm'** backend, and the model is prepared for quantization using **torch.quantization.prepare().** Subsequently, a new model **(quantized_model)** is formed as a sequence of quantization stubs, the original model, and dequantization stubs. The quantized model is then evaluated on the test dataset, and its accuracy is calculated and printed, providing insights into its performance on the given test data.

$$Q(x) = Round(max(x) - min(x)x - min(x) \times (quant_range - 1)) + min_quant_value \quad \text{(vi)}$$

Q(x) is the quantized value of input $x$
min(x) is the minimum value in the input tensor $x$
max(x) is the maximum value in the input tensor $x$
quant_rangequant_range is the range of quantization (i.e., for 8-bit quantization)
min_quant_value is the minimum quantized value.

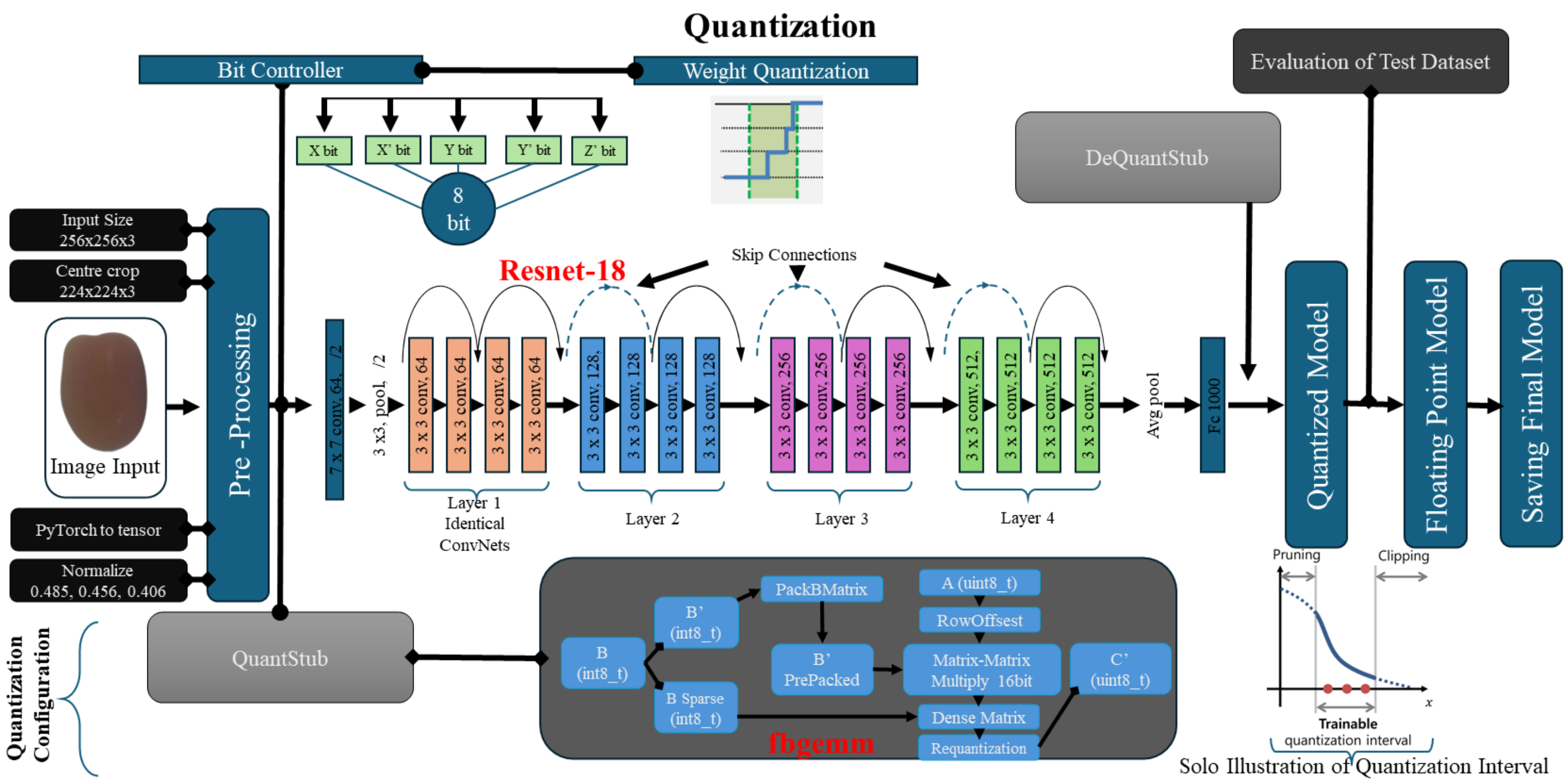

Figure 8: Implementation of quantization applied to ResNet-18 architecture.

### 2.5 Training

The total dataset was divided into training set and a test set, each containing images of tomatoes split into 90:10 (270:30 images) respectively. The validation dataset was further split from the training dataset i.e., 27 images. Importantly, to prevent data leakage, the division of the dataset was conducted by tomato sample. This means that for each unique tomato, all images were kept together in one partition, ensuring that no images from the same sample appear in both the training and test sets. The dataset is organized and loaded using the PyTorch DataLoader, facilitating efficient training and testing. Images undergo a sequence of preprocessing steps to enhance the model's ability to learn relevant features. The transformations include resizing images to 256x256 pixels, center cropping to 224x224 pixels, and normalizing pixel values. The normalization was achieved using the mean [0.485, 0.456, 0.406] and standard deviation [0.229, 0.224, 0.225] across the RGB channels. The training process was orchestrated using Stochastic Gradient Descent (SGD) as the optimizer, with a learning rate of 0.001 and momentum of 0.9. The criterion for optimization was the Cross-Entropy Loss. The training loop spans a total of 35 epochs, where the model was iteratively fine-tuned on the training dataset. During training, the model was set to training mode (resnet18.train()), and the training dataset was fed through the network in batches of size 4. The optimizer was used to minimize the computed loss through backpropagation. The training loss was calculated for each epoch, providing insights into the model's learning progress. To prevent overfitting, an early stopping mechanism is incorporated. If the average validation loss does not improve over a set number of consecutive epochs (5 in this study), training was halted to avoid unnecessary model refinement.

## 3. Performance Metrics

Various metrics were employed to assess the models' effectiveness, including classification accuracy, confusion matrix, precision, recall, F1-score, ROC curve, and the Precision-Recall Curves. Classification accuracy is a fundamental metric used to measure the predictive performance of a classification model. It is calculated by dividing the number of correct predictions by the total predictions, as illustrated by equation (vii).

$$Accuracy = \frac{No.\ of\ correct\ predictions}{Total\ no.of\ predictions} \times 100 \quad \text{(vii)}$$

A confusion matrix was drawn to delve deeper into the model's predictive capabilities. This matrix provided a detailed breakdown of the alignment between the model's predictions and the actual ground-truth labels for each class. While not a performance metric itself, the confusion matrix forms the basis for deriving other crucial metrics like precision, recall, and F1-score. Precision evaluated the accuracy of positive predictions, ensuring that positive predictions are reliable and do not falsely classify tomatoes. Equation (viii) depicts the precision calculation:

$$Precision = \frac{True\ Positives}{True\ Positives + False\ Positives} \quad \text{(viii)}$$

Recall, also known as sensitivity or true positive rate, measured the model's capacity to avoid missing any true maturity class of tomatoes. Equation (ix) represents the recall calculation:

$$Recall = \frac{True\ Positives}{True\ Positives + True\ Negatives} \quad \text{(ix)}$$

The F1-score combines precision and recall into a single metric, emphasizing a balance between correctly identifying tomatoes and not missing any of them. The harmonic mean is employed for this calculation, as shown in equation (x).

$$F_1 = \frac{2}{\frac{1}{precision} + \frac{1}{recall}} \quad \text{(x)}$$

The AUC-ROC curve visually represents a classification model's discriminatory power. It plots the True Positive Rate (sensitivity) against the False Positive Rate (1-specificity) across different classification thresholds. In addition to AUC-ROC, Precision-recall curve offers the advantage of comprehensively evaluating a classifier's performance across all threshold levels by plotting the True Positive Rate against the False Positive Rate. It is insensitive to class imbalance and allows for the calculation of the Area Under the Curve (AUC), which summarizes the model's ability to distinguish between classes, making it a powerful tool for comparing and selecting optimal models.

## 4. Results and Discussion

Table 3 and Figure 9 show the Performance metrics (Accuracy, Precision, Recall, and F1-Score) and confusion metrics for the ResNet model applied in detecting maturity indices of tomatoes, respectively. The evaluation of the three models as transfer learning, pruned + auto-tuned, and quantized showed differences in their ability to classify tomato maturity stages accurately. The quantized model achieved the overall accuracy of 97.81%, slightly surpassing the transfer learning model's accuracy of 97.78%, while the pruned + auto-tuned model has little bit lower with an accuracy of 87.96%. The quantized model demonstrated slightly better performance across all metrics with an overall accuracy of 97.81%, precision of 97.00%, recall of 97.20%, and an f1-score of 97.10% compared to transfer learning model. The quantization process reduced the model's computational complexity by using lower-precision representations yet effectively preserved critical feature representations necessary for accurate classification. Since the quantized model originated from the ResNet-18 model, its slightly better performance was explained by reduced noise in computations due to the lower precision, which acted as a form of regularization, preventing overfitting. The model achieved 100%) accuracy in the green, light red, red, and over-mature stages. In the more challenging breaker and pink stages, it maintained a high accuracy of 93.33% each, with minimal misclassifications, primarily in the green stage. The quantization process compresses the model by using lower-precision data types, enhancing efficiency and making it suitable for real-time applications in resource-constrained environments, such as agricultural fields. The high precision and recall values indicate that the model consistently identifies true positives while minimizing false positives across all maturity stages. The transfer learning model, while closely following the quantized model, had a slightly lower overall accuracy of 97.78%, precision of 96.50%, recall of 96.80%, and an f1-score of 96.65%. The model also achieved 100% accuracy in the green, light red, red, and over-mature stages. However, in the breaker and pink stages, the accuracy was slightly lower at 93.33%, with misclassifications in the green stage. The minimal difference in performance compared to the quantized model becomes significant when considering computational efficiency for practical deployment. The transfer learning model compared to quantized model does not offer the same computational benefits. Transfer learning models, derived from pre-trained models on large datasets, offer a substantial number of parameters that continue to require higher computational resources and memory, limiting efficiency gains. This model remains a viable option in environments with abundant computational resources due to its strong performance. However, it is less advantageous for applications where efficiency is paramount than the quantized model. The Pruned + auto-tuned model exhibited the lowest performance, with an overall accuracy of 87.96%, precision of 84.30%, recall of 85.00%, and an f1-score of 84.65%. Despite achieving 100% accuracy in the light red, red, and over-mature stages, its performance in the breaker, pink, and green stages was considerably lower. The accuracies were 66.67% for breaker, 73.33% for pink, and 90% for green. The pruning process, intended to eliminate redundant parameters, inadvertently removed essential features, leading to increased misclassifications and a decline in both precision and recall. The pruned + auto-tuned model highlights the risks associated with aggressive model optimization techniques. Pruning reduced the model's ability to distinguish between maturity stages with subtle differences, such as breaker and pink stages. The significant drop in accuracy, precision, recall, and f1-score indicate that essential features were removed, compromising the model's generalization capability. This underperformance renders it less suitable for tasks where accurate classification is critical.

Table 3: Performance metrics (Accuracy, Precision, Recall, and F1-Score) for ResNet model applied in detecting maturity indices of tomato.

| Model | Accuracy (%) | Precision (%) | Recall (%) | F1-Score (%) |
|---|---|---|---|---|
| Transfer Learning | 97.78 | 96.50 | 96.80 | 96.65 |
| Pruned + Auto-Tuned | 87.96 | 84.30 | 85.00 | 84.65 |
| Quantized | 97.81 | 97.00 | 97.20 | 97.10 |

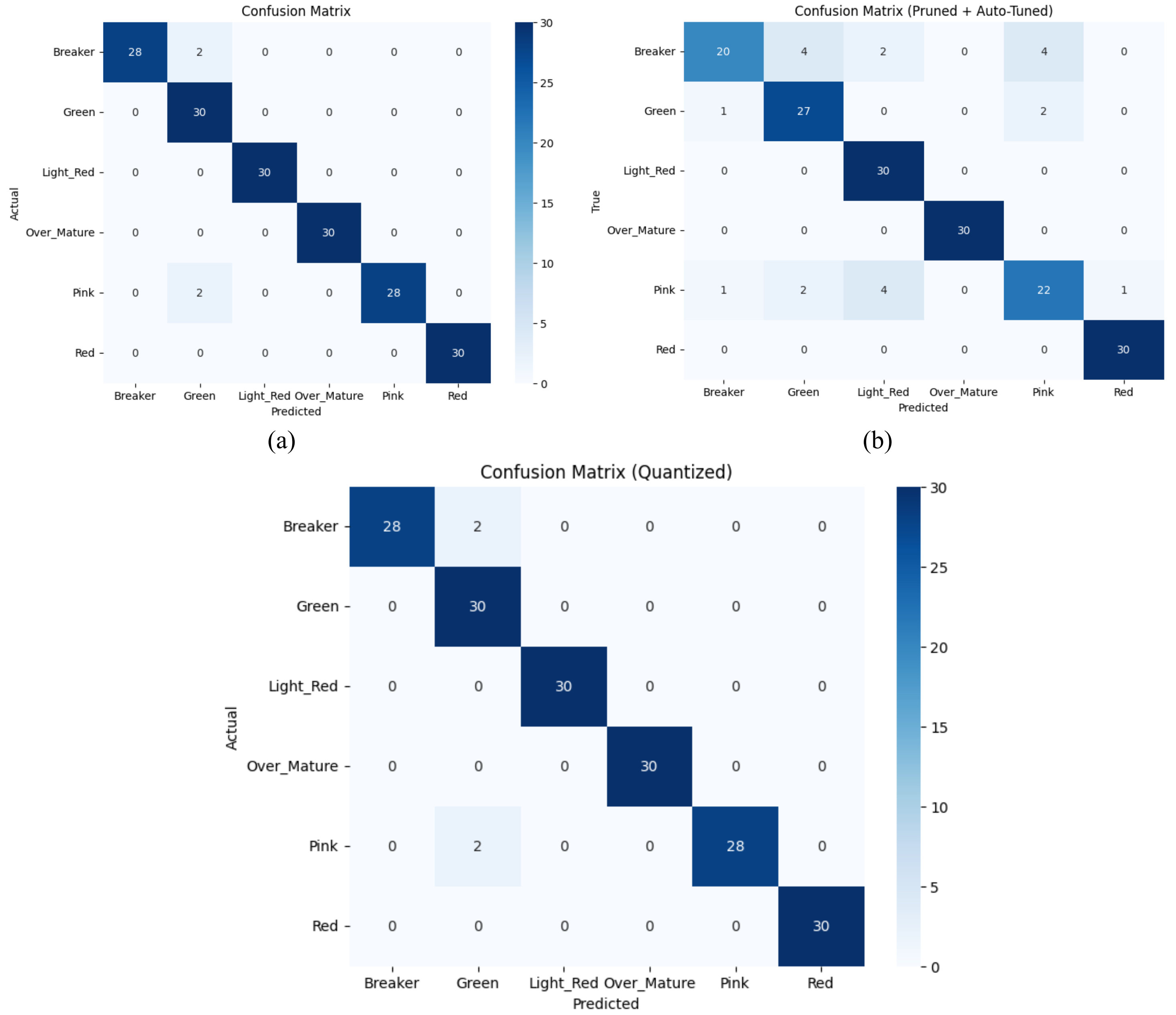

(a) (b)

(c)

Figure 9: Confusion metrics for ResNet model representing predicted classes against true classes used for detecting maturity indices of tomato; (a) Transfer Learning model, (b) Pruned + Auto tuned model, (c) Quantized model.

The ROC curve analysis revealed that the quantized model consistently outperformed the other models across all tomato maturity stages. For the quantized model, the breaker and pink stages achieved an AUC of 0.99, indicating near-perfect sensitivity and specificity. The green, light red, red, and over-mature stages each achieved a perfect AUC of 1.00, reflecting the model's ability to distinguish these stages without error. This performance suggests that the quantization process effectively preserved the model's discriminative power, enabling it to maintain high true positive rates while minimizing false positives. The PR curve analysis of the quantized model showed that the model achieved high average precision scores across all maturity stages, with scores of 1.00 for the light red, red, and over-mature stages. For the breaker and pink stages, the average precision was 0.95 and 0.93, respectively. These high precision levels, even at increased recall values, indicate the model's robustness in correctly identifying true positives while minimizing false positives. The stable and high PR curves suggest that the quantized model effectively handles the trade-off between precision and recall, maintaining consistent performance across different thresholds. The complete illustration of ROC and PR curves are given in Figure 10 and 11 respectively.

In comparison, the transfer learning model also demonstrated strong performance, with AUCs of 0.99 for the breaker and pink stages and 1.00 for the remaining stages. The pruned + auto-tuned model, however, exhibited diminished performance, particularly in the breaker and pink stages, with AUCs of 0.93 and 0.95, respectively. These lower AUC values indicate challenges in the model's ability to maintain low false positive rates, likely due to the loss of critical discriminative features during the pruning process. The transfer learning model exhibited similar trends in the PR curves, with average precision scores of 0.95 for the breaker stage and 0.93 for the pink stage. Although these scores are commendable, the slight edge of the quantized model highlights its enhanced ability to retain critical feature representations after quantization. The pruned + auto-tuned model showed significantly lower average precision scores of 0.79 for the breaker stage and 0.80 for the pink stage. The substantial drop in precision as recall increased indicates a higher rate of false positives, reflecting the model's struggle to distinguish between stages with subtle differences due to the pruning process.

The analyses of both ROC and PR curves underscore the quantized model's ability to maintain high classification performance while reducing computational complexity. The quantization process not only compressed the model but also preserved essential discriminative features necessary for accurate classification. This is particularly advantageous in agricultural applications where computational resources are limited, and rapid, precise classification is essential for tasks such as automated harvesting and sorting. In contrast, the diminished performance of the pruned + auto-tuned model highlights the potential drawbacks of aggressive model optimization techniques. The pruning inadvertently eliminated important features crucial for distinguishing between maturity stages with subtle visual differences, leading to decreased sensitivity and precision, especially in the breaker and pink stages. This outcome emphasizes the importance of carefully balancing model simplification with the preservation of critical feature representations to ensure reliable performance.

The ability of quantized models to accurately classify all tomato maturity stages, combined with reduced model size and increased inference speed, makes it suitable for integration into edge devices platforms. The quantized model's performance demonstrates that quantization is an effective strategy for optimizing deep learning models for real-world applications without sacrificing accuracy. Future research could explore further enhancements to the quantization process and investigate hybrid optimization techniques that combine quantization with selective pruning to achieve even greater efficiency while maintaining high classification performance.

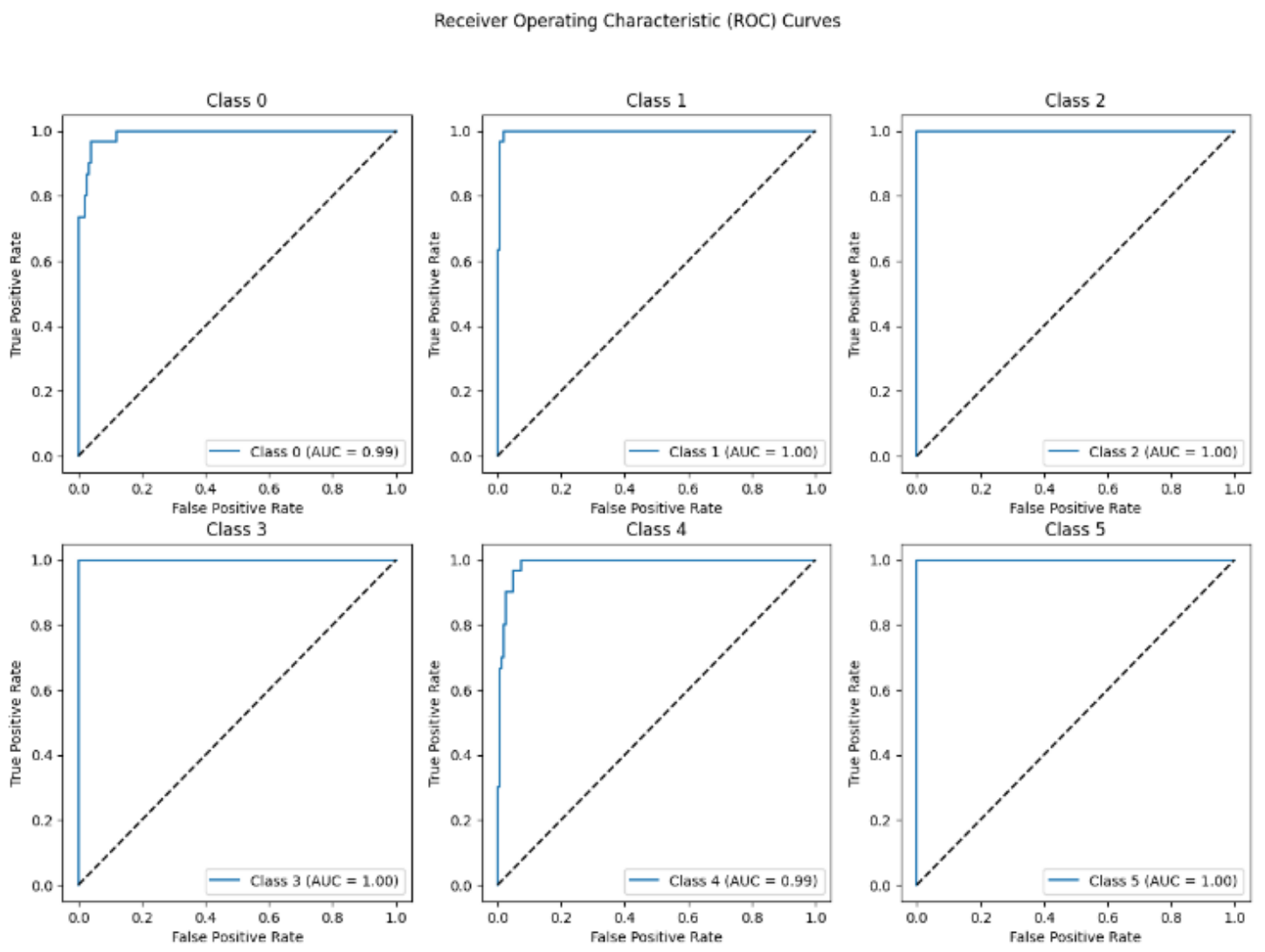

(a)

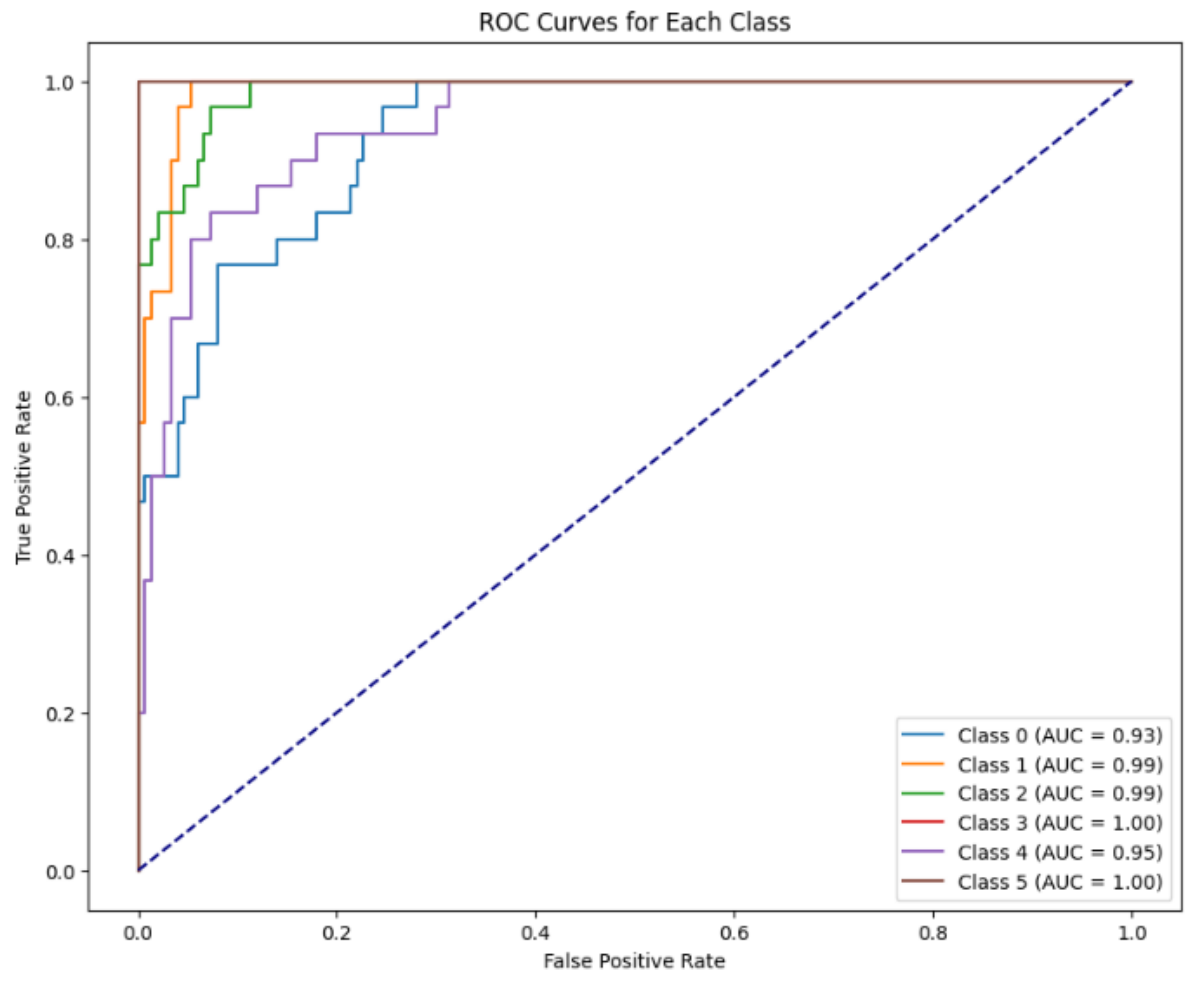

(b)

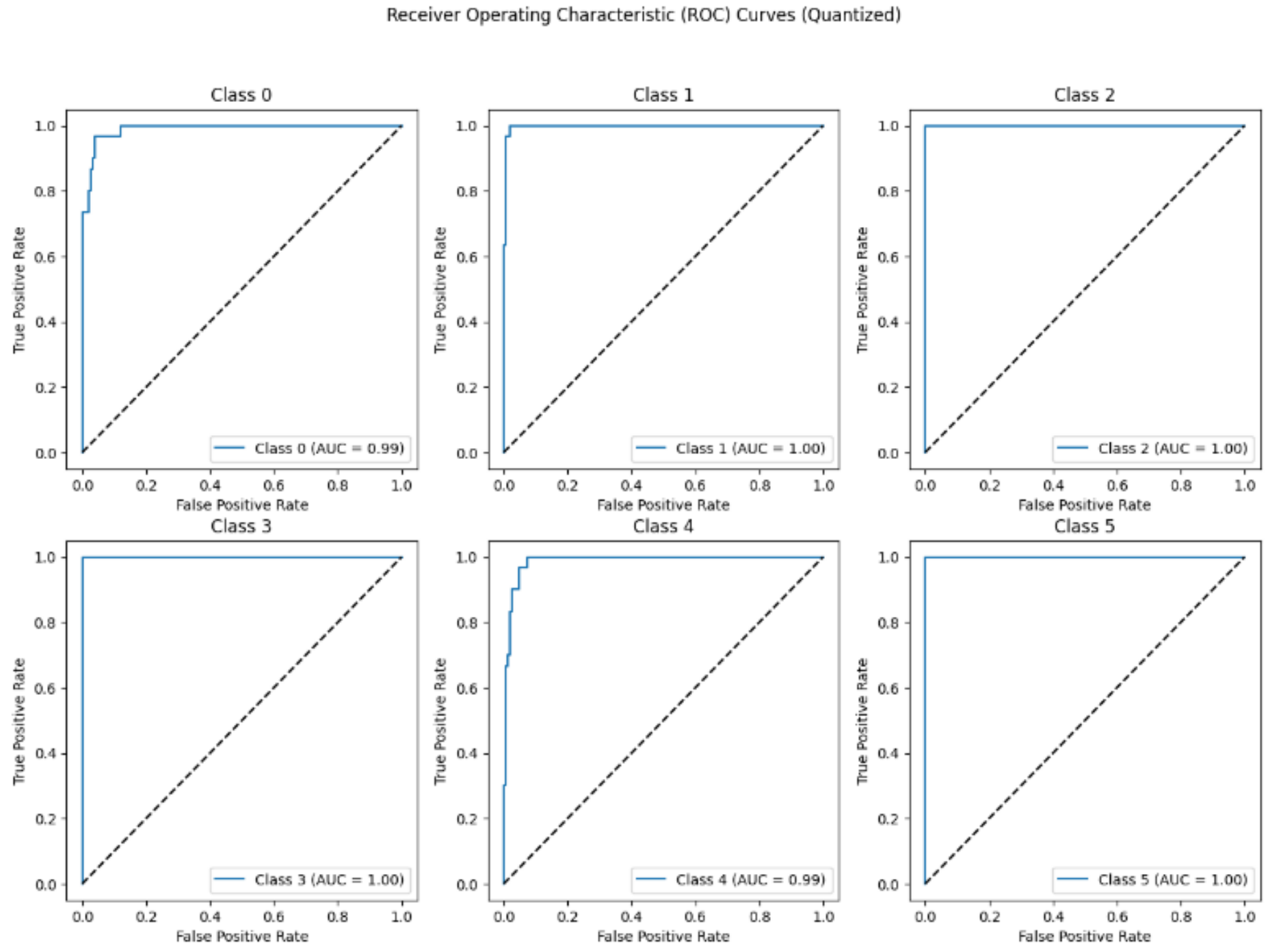

(c)

Figure 10: ROC (Receiver Operating Characteristic) curves illustrating the sensitivity (true positive rate) versus the specificity (false positive rate) of the ResNet model (a) Transfer Learning model, (b) Pruned + Auto tuned model, (c) Quantized model used for detecting tomato maturity indices.

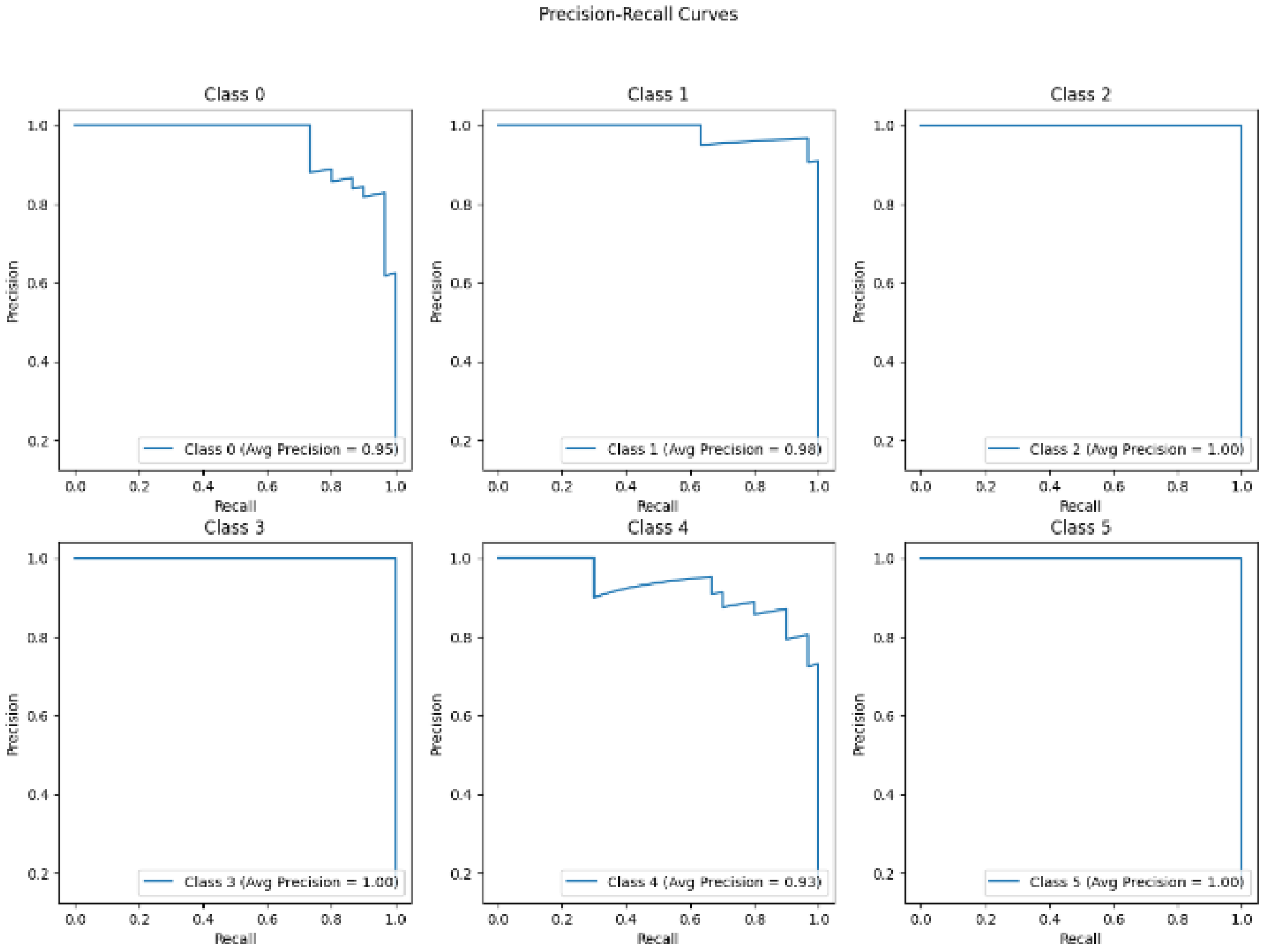
Precision-Recall Curves
Class 0
Class 0 (Avg Precision = 0.95)
Class 1
Class 1 (Avg Precision = 0.98)
Class 2
Class 2 (Avg Precision = 1.00)
Class 3
Class 3 (Avg Precision = 1.00)
Class 4
Class 4 (Avg Precision = 0.93)
Class 5
Class 5 (Avg Precision = 1.00)
Precision
Recall

(a)

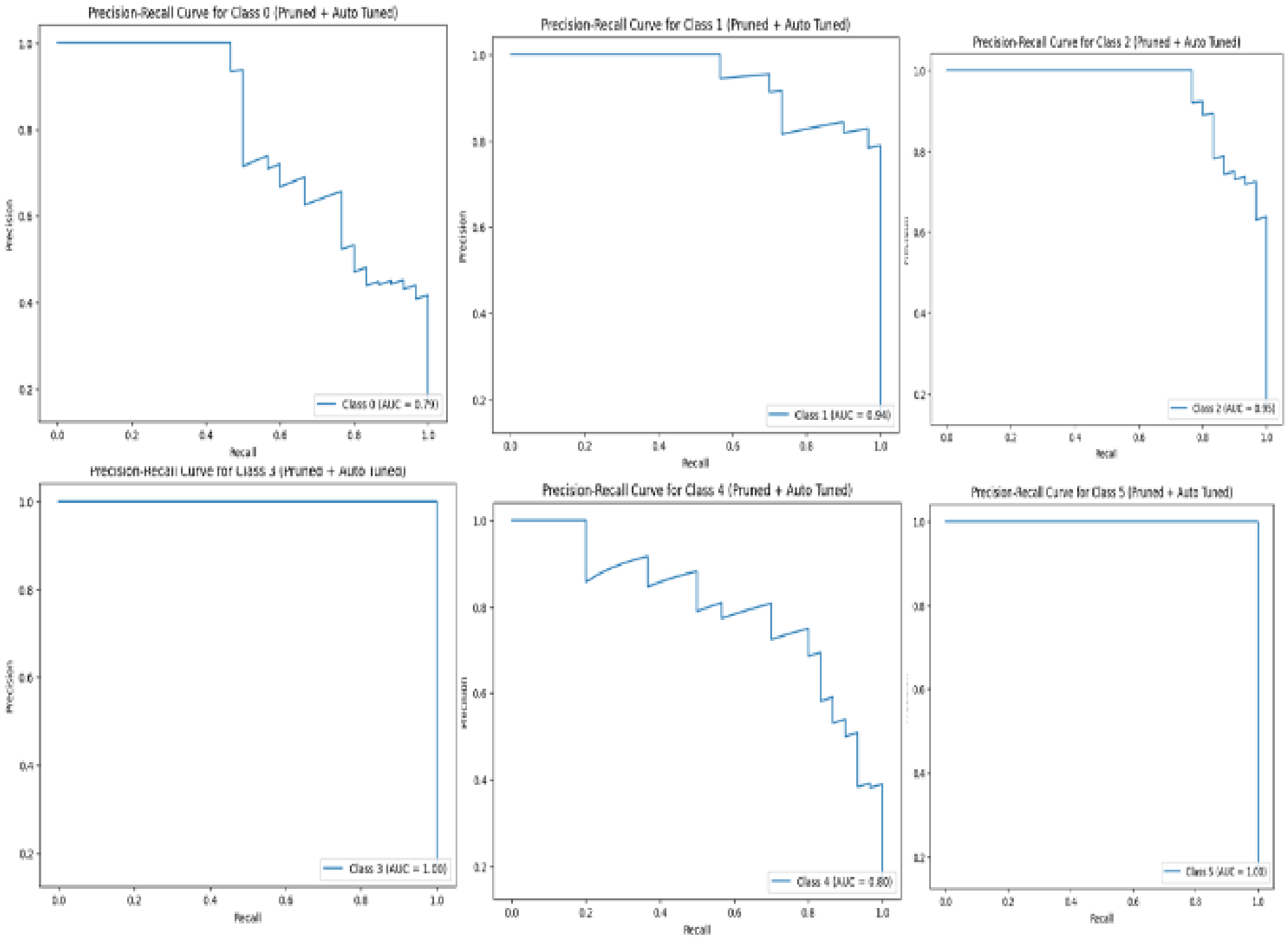
Precision-Recall Curve for Class 0 (Pruned + Auto Tuned)
Class 0 (AUC = 0.79)
Precision-Recall Curve for Class 1 (Pruned + Auto Tuned)
Class 1 (AUC = 0.94)
Precision-Recall Curve for Class 2 (Pruned + Auto Tuned)
Precision-Recall Curve for Class 3 (Pruned + Auto Tuned)
Class 3 (AUC = 1.00)
Precision-Recall Curve for Class 4 (Pruned + Auto Tuned)
Class 4 (AUC = 0.80)
Precision-Recall Curve for Class 5 (Pruned + Auto Tuned)
Class 5 (AUC = 1.00)
Precision
Recall

(b)

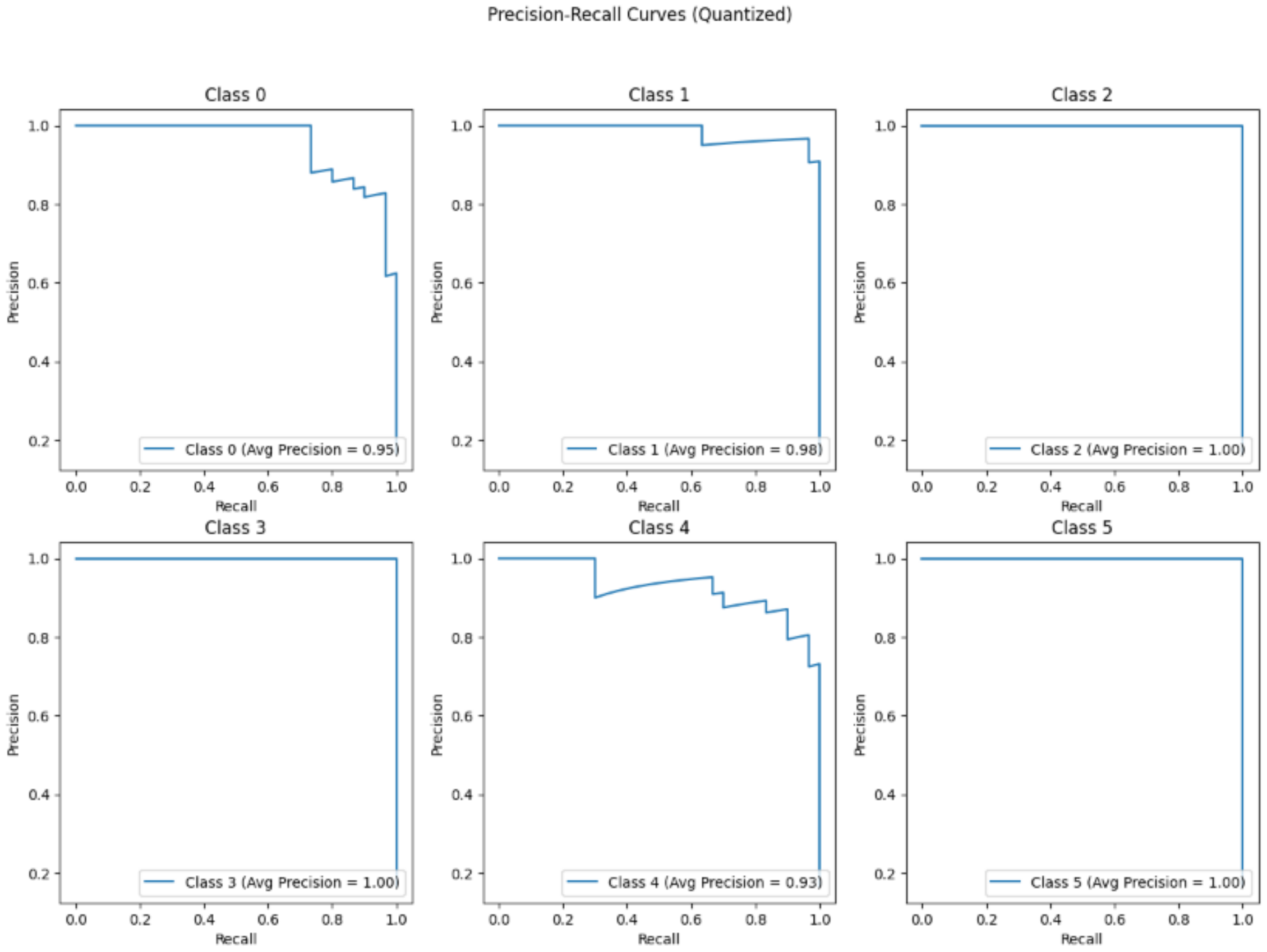

(c)

Figure 11: Precision, recall curve of the ResNet model (a) Transfer Learning model, (b) Pruned + Auto tuned model, (c) Quantized model used for detecting tomato maturity indices.

## 4.1 Deployment of Models on Jetson Nano

After the models' development and validation stages, their deployment on the jetson nano platform provides crucial insights into their operational efficiency and performance under real-world conditions. This section deals with the specifics of deploying the transfer learning, pruned + auto-tuned, and quantized models on the Jetson Nano, evaluating key metrics such as build time, model load time, CPU and GPU usage, maximum processing unit temperature, and average classification time as illustrated in Table 4.

The transfer learning model exhibited the longest build time of 3 minutes and 40 seconds and a model load time of 74 seconds, reflecting its complexity and larger size. Its average CPU and GPU usage during inference was high at 80%, and it reached a maximum temperature of 54°C. The average classification time was the slowest among the models at 0.01201 seconds per image. These results indicate that while the transfer learning model is accurate, its substantial computational demands make it less suitable for deployment on resource-constrained devices like the jetson nano. In contrast, the pruned + auto-tuned model demonstrated improved deployment efficiency with a reduced build time of 1 minute and 22 seconds and a model load time of 40 seconds. The CPU and GPU usage decreased to 70%, and the maximum temperature lowered to 52°C. Notably, the average classification time significantly improved to 0.001073 seconds per image. However, this efficiency gain came at the cost of reduced accuracy and precision in certain maturity stages, suggesting that critical features may have been lost during the pruning process. This trade-off highlights the challenges of aggressive model optimization techniques that may compromise essential predictive capabilities. The quantized model achieved the best balance between performance and efficiency. With a build time of 1 minute and 56 seconds and a model load time of 56 seconds, it offered moderate initialization periods. The model's CPU and GPU usage was the lowest at 65%, and it maintained the lowest maximum temperature of 51°C, indicating optimal resource utilization. Additionally, it achieved the fastest average classification time of 0.000975 seconds per image while maintaining high accuracy across all maturity stages. The reduced computational complexity translates to faster inference times and lower power consumption, which are critical factors for deployment on edge devices and mobile platforms. These findings demonstrate that quantization effectively reduces computational demands without sacrificing model performance, making it suitable for deployment on edge devices like the jetson nano.

The deployment results have significant implications for the practical application of deep learning models in resource-constrained environments. The quantized model's better operational efficiency and rapid inference times make it ideal for real-world agricultural applications where quick and accurate classification is critical, such as automated harvesting and sorting systems. Its ability to maintain accuracy with reduced computational requirements suggests that quantization is an effective strategy for optimizing models for edge deployment. The observed drawbacks of the pruned + auto-tuned model emphasize the need for cautious application of pruning techniques. While pruning can enhance efficiency, it may inadvertently remove essential features necessary for accurate classification, particularly in stages with subtle differences. Future research should focus on developing advanced pruning methods that preserve critical features or exploring hybrid optimization approaches that combine pruning with quantization to balance efficiency and accuracy. Moreover, real-world testing of the quantized model on the jetson nano in operational agricultural settings would provide valuable insights into its performance under practical conditions. Such evaluations could inform further refinements to the model and optimization processes, ensuring robustness and reliability in diverse environments. Expanding the scope of deployment studies to include other edge devices and exploring the impact of varying environmental factors on model performance would also be beneficial.

Table 4: Deployment Metrics of Models on Jetson Nano

| Metric | Transfer Learning | Pruned + Auto-Tuned | Quantized |
|---|---|---|---|
| Build Time (minutes:seconds) | 3:40 | 1:22 | 1:56 |
| Model Load Time (seconds) | 74 | 40 | 56 |
| CPU and GPU Usage (%) | 80 | 70 | 65 |
| Maximum Temperature (°C) | 54 | 52 | 51 |
| Average Classification Time (s/image) | 0.01201 | 0.001073 | 0.000975 |

This study employed a quantized model for tomato maturity detection under controlled conditions, demonstrating excellent performance. However, for real-world agricultural applications, future studies must address key challenges related to environmental variability. A diverse dataset incorporating fluctuating lighting, varying backgrounds, and other real-world complexities will be critical to ensure accurate and robust performance in uncontrolled environments. Deep learning models trained on such diverse datasets will better reflect the conditions encountered in field applications. Expanding this work to different crop varieties and physiological traits is another area for future exploration. The ability of the quantized model to operate on low-power edge devices with minimal resource consumption paves the way for widespread adoption, particularly among small and medium-sized farms. By providing efficient, real-time analysis of crop maturity, the results from this study could be used for timely decisions regarding harvest timing and disease management, ultimately improving yield and reducing crop losses. Moreover, the findings from this study could be scaled up for use in large-scale farming operations, facilitating automated sorting and grading of agricultural produce through industrial conveyors and on-farm systems in warehouses and processing facilities. As agricultural systems become increasingly data-driven, cutting-edge computing and AI will be indispensable in providing real-time insights directly from the field. However, issues such as data privacy and security will become critical beyond performance and scalability. Since edge devices process data locally, this approach enhances data security by reducing transmission to centralized servers. In addition, lightweight models optimized for edge devices can further streamline data collection and analysis processes, minimizing cloud-based system dependency that requires more resources. The comparative analysis of recent existing studies with ours is illustrated in table 5.

Table 5. Comparative analysis of recent studies

| Model | Approach | Accuracy % | Draw back | Reference |
|---|---|---|---|---|
| Quantized ResNet | Our Study | 97.81 | - | - |
| Transfer Learning | Our Study | 97.78 | High computational resources required | |
| Pruned + Auto-Tuned ResNet | Our Study | 87.96 | Loss of essential features due to aggressive pruning | |
| YOLOv5s-tomato | Improved YOLOv5 Model | 97.42 | Struggle with small-target tomatoes | (R. Li et al., 2023) |
| MHSA-YOLOv8 | MHSA-YOLOv8 Model | 86.42 | Increased model size due to MHSA integration | (P. Li et al., 2023) |
| MTS-YOLO | MTS-YOLO Model | 92.0 | Require optimization for deployment | (Wu et al., 2024) |
| SE-YOLOv3-MobileNetV1 | SE-YOLOv3-MobileNetV1 Network | 95.0 | Degrade in natural greenhouse environments | (Su et al., 2022) |
| EfficientNetB0 | Transfer Learning Approach | 90.0 | Require large datasets for optimal performance | (Lumoring et al., 2024) |
| YOLOv9 | YOLOv9 Model | 92.49 | Potential overfitting | (Vo et al., 2024) |

## 5. Limitations and Future directions:

While our proposed tomato maturity classification model demonstrates promise in efficiently classifying maturity stages in low-power edge devices, few limitations are there. Our study utilized 25 unique tomato samples per maturity stage, and although capturing multiple perspectives helped enrich the feature set, this relatively small number of unique samples may constrain the generalizability of the results; future work should involve larger and more diverse datasets to fully capture the variability encountered in real-world agricultural settings. Moreover, the controlled image acquisition conditions characterized by consistent lighting and uniform backgrounds do not fully reflect the environmental challenges such as variable illumination, diverse backgrounds, or occlusions that are typical in field conditions, potentially impacting the model's performance when deployed in field instead of industry. Furthermore, even though our optimized model shows real-time performance on a specific low-power edge device, the deployment was tested under limited hardware configurations and evaluation scenarios, meaning that variations in hardware capabilities or integration challenges could alter its effectiveness across different agricultural settings. Finally, while quantization and pruning techniques significantly reduced computational demands without a substantial loss in accuracy, these optimization methods might introduce trade-offs that affect the model's ability to discriminate subtle differences in maturity levels under challenging conditions, indicating a need for further research to understand their long-term impact on model reliability across broader application scenarios. Additionally, investigating hybrid optimization techniques such as combining model quantization with selective pruning could enhance efficiency while maintaining critical features for accurate classification. Extending the quantized model framework to other crops and maturity indices offers a promising avenue to assess its versatility across various agricultural products. Finally, the development of user-friendly deployment frameworks would facilitate seamless integration of the quantized model into existing agricultural workflows, increasing its accessibility for farmers and agricultural professionals.

## 6. Conclusion

The primary goal of this research was to develop a computationally efficient tomato classification model using ResNet-18 architecture. This study seeks to overcome the challenges posed by computationally intensive deep learning networks using pruning and quantization approaches, enabling their deployment in resource-constrained agricultural environments. The quantized model emerged as the suitable option for deployment on the Jetson Nano, achieving an accuracy of 97.81% and the lowest classification time of 0.000975 seconds per image. It also demonstrated the efficient resource usage, with 65% CPU and GPU utilization, making it suitable for real-world applications in agriculture. The pruned and auto-tuned model showed improved deployment efficiency but required careful balancing to maintain performance, with an accuracy of 87.96%. These findings emphasize the importance of optimizing both computational efficiency and accuracy for edge device deployment. Future work will focus on extending these techniques to other crops and agricultural processes to enhance productivity and sustainability in diverse environments.

**Conflict of Interest:**

All authors have no conflict of interest.

**Acknowledgement:**

This research was supported by National Research Program for Universities, Higher Education Commision (HEC), Pakistan (20-15545/NRPU/RandD/HEC/2021 2021) and University of Wyoming, USA.

**Author Contribution Statement:**

**Muhammad Waseem**: Conceptualization; Methodology; Writing; Experiment design, **Chung-Hsuan Huang**: Conceptualization; Methodology, **Muhammad Muzzammil Sajjad**: Analysis; Validation; Writing, **Laraib Haider Naqvi**: Data curation; Visualization; Editing, **Yaqoob Majeed**: Administration; Supervision; Correspondence; Reviewing & Editing, **Tanzeel Ur Rehman**: Software; Validation; Investigation; Reviewing & Editing, **Tayyaba Nadeem**[5]: Resources; Editing